\documentclass[]{bytedance}
\usepackage[toc,page,header]{appendix}

\usepackage{minitoc}
\usepackage{amsfonts}
\usepackage{amssymb}
\usepackage{tabularx}
\usepackage{listings}
\usepackage{xcolor}
\usepackage{cancel}

\usepackage{tabulary,multirow,xspace}
\usepackage{fixmath,mathtools,nicefrac,mmstyle}
\usepackage{subcaption}
\usepackage{caption}
\usepackage{wrapfig} 
\usepackage[misc]{ifsym} 
\usepackage{colortbl}

\usepackage{wrapfig}
\usepackage{multicol}
\usepackage[most]{tcolorbox}
\usepackage{pifont}

\definecolor{codegreen}{rgb}{0,0.6,0}
\definecolor{codegray}{rgb}{0.5,0.5,0.5}
\definecolor{codepurple}{rgb}{0.58,0,0.82}
\definecolor{backcolour}{rgb}{0.95,0.95,0.92}
\definecolor{boxblue}{RGB}{57,89,163}
\definecolor{boxbluebg}{RGB}{230,237,250} 

\lstdefinestyle{mystyle}{
    backgroundcolor=\color{backcolour},   
    commentstyle=\color{codegreen},
    keywordstyle=\color{magenta},
    numberstyle=\tiny\color{codegray},
    stringstyle=\color{codepurple},
    basicstyle=\ttfamily\footnotesize,
    breakatwhitespace=false,         
    breaklines=true,                 
    captionpos=b,                    
    keepspaces=true,                 
    numbers=none,                    
    numbersep=5pt,                  
    showspaces=false,                
    showstringspaces=false,
    showtabs=false,                  
    tabsize=2
}
\definecolor{mygray1}{gray}{.95}
\definecolor{mygray2}{gray}{.9}
\definecolor{mygray3}{gray}{.95}
\usepackage{pifont}

\newlength\savewidth
\newcolumntype{x}[1]{>{\centering\arraybackslash}p{#1pt}}

\newcommand{\app}{\raise.17ex\hbox{$\scriptstyle\sim$}}

\usepackage{xcolor}
\usepackage{graphicx}
\usepackage{amssymb}
\usepackage{pifont}
\usepackage{floatrow}
\usepackage{amsmath} 
\usepackage{float}
\usepackage{wrapfig}
\usepackage{multirow}
\usepackage{tcolorbox}
\tcbuselibrary{breakable, skins, raster}
\usepackage{listings}
\usepackage{listings}

\usepackage{algorithm}
\usepackage{algorithmic}
\definecolor{myblue}{RGB}{210, 225, 255}
\definecolor{mytextblue}{RGB}{51, 161, 201}
\definecolor{mypurple}{RGB}{218, 112, 214}

\definecolor{commentgreen}{rgb}{0.1, 0.4, 0.1}
\definecolor{keywordblue}{rgb}{0.1, 0.1, 0.7}
\definecolor{stringred}{rgb}{0.7, 0.1, 0.1}

\lstdefinestyle{mystyle}{
    commentstyle=\color{commentgreen},
    keywordstyle=\color{keywordblue},   
    stringstyle=\color{stringred},
    basicstyle=\ttfamily\scriptsize, 
    breaklines=true,
    keepspaces=true,
    showstringspaces=false,
    frame=none,                     
    language=Python, 
}

\usepackage{dsfont}

\newcommand{\name}{UMO}
\title{\name{}: Scaling Multi-Identity Consistency for Image Customization via Matching Reward}

\author{
    \centerline{
        Yufeng Cheng \quad
        Wenxu Wu \quad
        Shaojin Wu \quad
        Mengqi Huang $^*$ \quad
    }
    \centerline{
        Fei Ding $^{\dagger}$ \quad
        Qian He \quad
    }
}

\affiliation[]{UXO Team, Intelligent Creation Lab, ByteDance}

\footnotetext{* Corresponding author $\dagger$ Project lead.}

\abstract{
       Recent advancements in image customization exhibit a wide range of application prospects due to stronger customization capabilities. However, since we humans are more sensitive to faces, a significant challenge remains in preserving consistent identity while avoiding identity confusion with multi-reference images, limiting the identity scalability of customization models. To address this, we present \textbf{\textit{UMO}}, a \textbf{U}nified \textbf{M}ulti-identity \textbf{O}ptimization framework, designed to maintain high-fidelity identity preservation and alleviate identity confusion with scalability. With multi-to-multi matching paradigm, UMO reformulates multi-identity generation as a global assignment optimization problem and unleashes multi-identity consistency for existing image customization methods generally through reinforcement learning on diffusion models. To facilitate the training of UMO, we develop a scalable customization dataset with multi-reference images, consisting of both synthesised and real parts. Additionally, we propose a new metric to measure identity confusion. Extensive experiments demonstrate that UMO not only improves identity consistency significantly, but also reduces identity confusion on several image customization methods, setting a new state-of-the-art among open-source methods along the dimension of identity preserving. Code and model: \url{https://github.com/bytedance/UMO}
}

\date{\today}

\checkdata[Project Page]{\url{https://bytedance.github.io/UMO}}

\correspondence{Yufeng Cheng at \email{chengyufeng.cb1@bytedance.com}}

\begin{document}
\maketitle

\section{Introduction}
\label{sec:intro}


Image customization, which aims to create images that simultaneously adhere to the semantic content of textual prompts and the visual appearance of reference images, has garnered significant research attention in recent years. Among various subjects, the customization of human identities (\textit{i.e.}, ID) has attracted particular interest due to its broad range of real-world applications, such as personalized film production and virtual avatar creation. Different from other subjects, humans are exceptionally sensitive to identity customization, \textit{i.e.}, even subtle discrepancies in appearance can lead to a noticeable loss of identity fidelity, thereby making human ID customization significantly more challenging. This challenge is further amplified when multiple identities need to be customized simultaneously, as it requires the model not only to preserve the unique characteristics of each individual ID, but also to maintain clear distinctions among them within the generated images.

\begin{figure}[H]
    \centering  
    \includegraphics[width=1\textwidth]{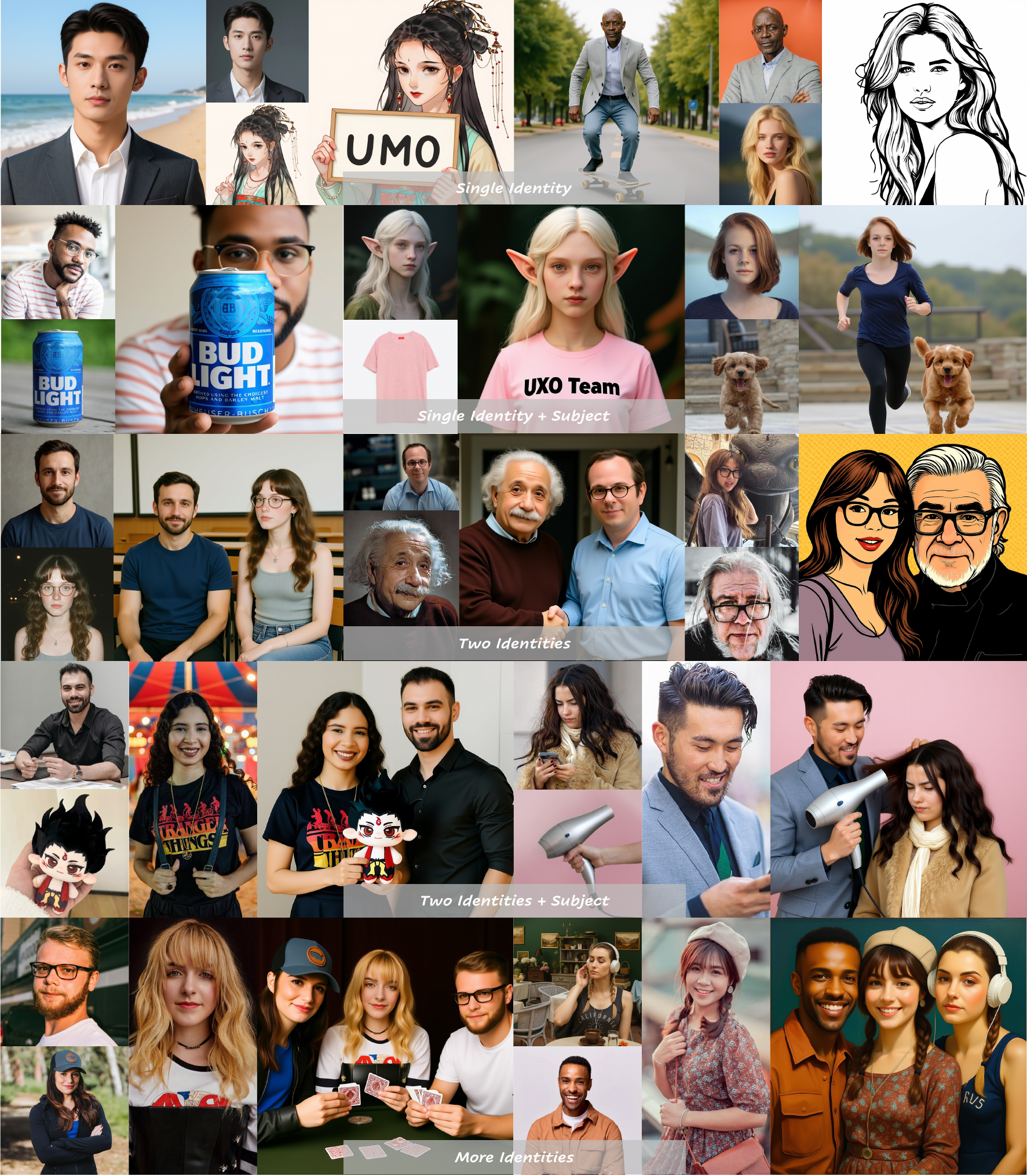}
    \caption{Showcase of our UMO model in different scenarios. The detailed prompts are listed in \Cref{tab:showcase_prompt}.}
    \label{fig:showcase}
\end{figure}

\begin{figure*}[t]
    \centering
    \includegraphics[width=\textwidth]{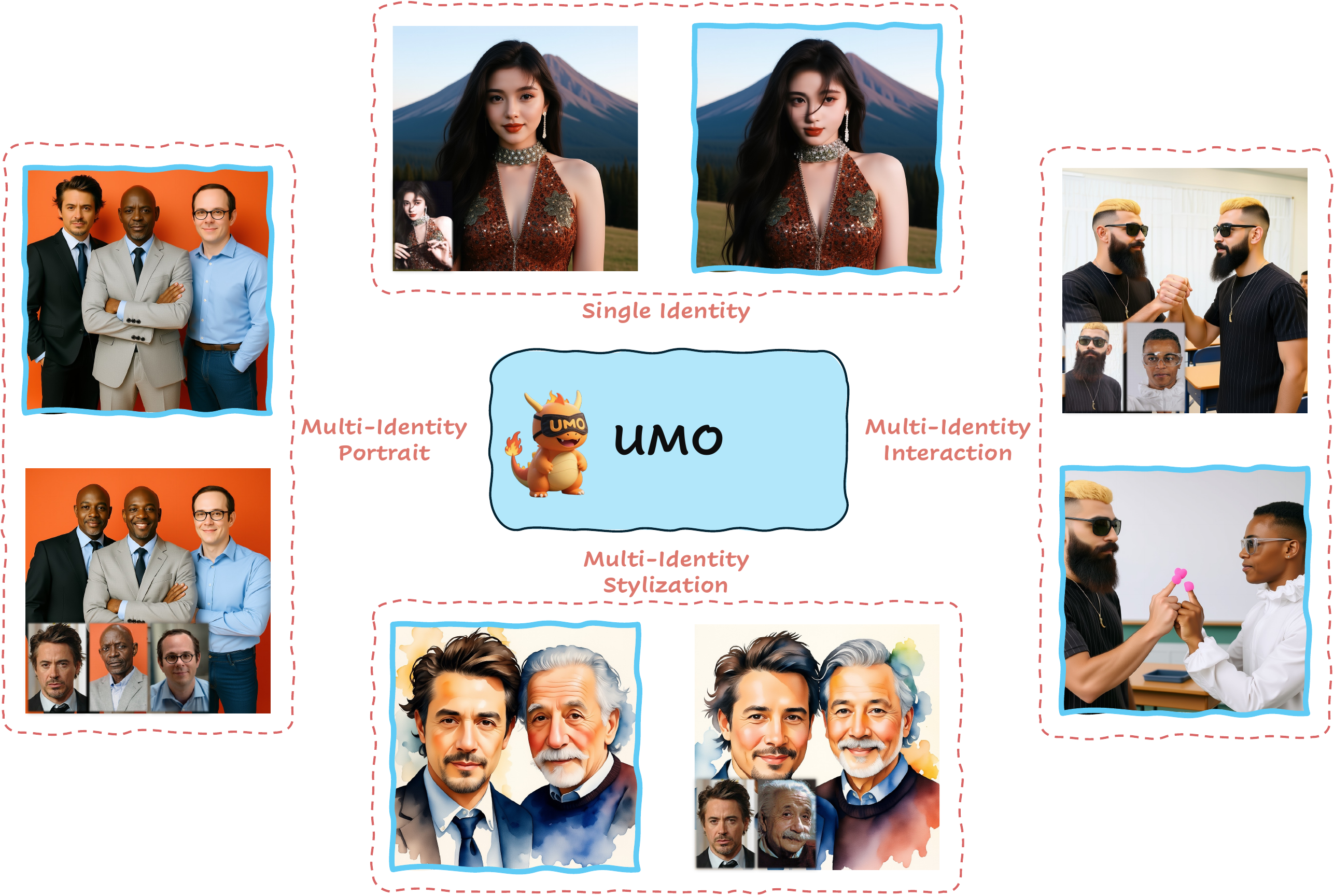}
    \caption{Our UMO unleashes multi-identity consistency and alleviates identity confusion. Existing image customization methods suffer low facial fidelity and severe identity confusion, while UMO can tackle these problems with results in \textbf{blue boxes}.}
    \label{fig:teaser}
\end{figure*}

Existing multi-identity customization methods primarily focus on constructing improved multi-identity paired data to enhance the consistency of multiple identities. For example, DreamO~\cite{dreamo} and OmniGen~\cite{omnigen} build large-scale training datasets for identity customization, including multi-identity paired data derived from either video sources or in-context image generation. Meanwhile, some recent approaches aim to mitigate confusion between multiple identities by employing identity masks to explicitly constrain the location or position of each identity. For instance, MS-Diffusion~\cite{msdiffusion} introduces a layout-guided mechanism to explicitly control the generation location of each identity. RealCustom++~\cite{realcustom, mao2024realcustom++} further proposes to explicitly separate the influence masks of each identity in order to disentangle their respective generations. In summary, existing methods mainly adopt a \textbf{one-to-one mapping} paradigm, in which focuses on learning a direct correspondence between each identity in the reference image and the corresponding generated one.

In this study, we argue that the existing \textit{one-to-one mapping} paradigm fails to comprehensively address both \textit{intra-ID variability} and \textit{inter-ID distinction}, leading to increased identity confusion and reduced identity similarity as the number of identities grows. On the one hand, \textit{intra-ID variability} refers to the inherent variability within a single identity, where the same individual may present different attributes (\textit{e.g.}, poses, expressions, \textit{etc.}) between the reference image and the generated output. On the other hand, \textit{inter-ID distinction} underscores the importance of not only accurately capturing the distinctive characteristics of the target identity during generation, but also explicitly suppressing the features associated with other identities, thereby ensuring clear separation and minimizing identity confusion in multi-identity scenarios. As the number of identities increases, the risk arises that the distinctions between different identities may become less salient, potentially approaching the degree of variability observed within a single identity. Therefore, by focusing exclusively on the \textit{one-to-one mapping} between each identity and its corresponding reference, existing methods overlook the increasing overlap between intra-ID variability and inter-ID distinction as the number of identities grows. This limitation fundamentally restricts their scalability, as they are unable to effectively preserve clear identity boundaries in large-scale multi-identity scenarios.


To address the challenge, we propose a novel \textbf{multi-to-multi matching} paradigm, whose key idea is to reformulate multi-identity generation as a global assignment optimization problem, aiming to maximize the overall matching quality between multiple identities and multiple references. Therefore, each generated identity can be paired with the most suitable reference, which maximizes inter-ID distinction while minimizing the impact of intra-ID variability, thereby enabling accurate and scalable multi-identity generation. Technically, to ensure that our method remains concise and readily applicable to various customized models to improve identity consistency, we propose a \textbf{U}nified \textbf{M}ulti-identity \textbf{O}ptimization (\textbf{UMO}) framework, which operationalizes the \textit{multi-to-multi matching} paradigm through a novel Reference Reward Feedback Learning (ReReFL). Specifically, UMO begins by defining a lightweight and reliable single-identity reference reward based on the cosine distance between identity embeddings. This is then scaled to a multi-identity context by formulating a bipartite graph between multiple references and generated identities, and optimizing the global matching score to achieve the optimal assignment. Moreover, to support the effective training of UMO, we develop a scalable customization dataset with multiple reference images for each identity, along with a new metric (ID-Conf) designed to precisely evaluate the extent of multi-identity confusion. 

Our main contributions can be summarized as follows:

\textbf{Concept: } We highlight that existing customization methods fails to address intra-ID variability and inter-ID distinction, as their \textit{one-to-one mapping} paradigm leads to decreased identity consistency with the scale of the identities. For the first time, we propose a novel \textit{multi-to-multi matching} paradigm that maximizes overall matching quality between multiple identities and references.

\textbf{Technique: } We propose UMO, a novel Reference Reward Feedback Learning framework featuring scalable multi-identity reference reward optimization, which can be easily integrated into various customization models.

\textbf{Metric: } We propose ID-Conf to accurately evaluate the extent of multi-identity confusion. For a reference identity, ID-Conf is defined as the relative margin between the two most similar generated candidates, given the observation that confusion occurs with several indistinct generated faces.

\textbf{Performance: } We conduct extensive experiments on XVerseBench~\cite{xverse} and OmniContext~\cite{omnigen2}, including multi-identity scenarios. Our UMO significantly boosts identity similarity and mitigates confusion simultaneously on various customized models, leading to the highest ID-Sim and ID-Conf scores. This demonstrates its strong and general promoting effect on identity consistency, showcasing its capability to deliver state-of-the-art~(SOTA) results with high fidelity identities as shown in \Cref{fig:showcase}.

\section{Related Works}
\label{sec:related}

\subsection{Multi-subject Driven Generation}

Text-to-image models~\cite{ldm, sdxl, sd3, flux, imagen, dalle3} experienced explosive growth in recent years. 
Despite the increasing text-instruction following capability of those models, they still struggle to fulfill a common requirement: generating images for given subjects with reference images.

Early works~\cite{dreambooth, textual_inversion} extended subject-driven generation from text-to-image (T2I) models through parameter-efficient fine-tuning on each given subject. Subsequent works~\cite{ipadapter, elite} achieved general single-subject injection by modifying and training components such as the cross-attention module.

Building on the general single-subject feature injection, numerous extensions have been proposed. A series of works, such as MIP-adapter~\cite{mip_adapter}, MSDiffusion~\cite{msdiffusion}, and UNO~\cite{uno} extend the method to general multi-subject driven generation by leveraging the flexible context window lengths of attention mechanisms. Another improvement aims at ID preservation. InstanceID~\cite{instantid}, Pulid~\cite{pulid} achieved higher ID preservation by adapting widely used face recognition models as encoders and training supervisions.
Other works directly train an image generation model from scratch to natively support both textual input and reference subject as input~\cite{omnigen, bagel, UniReal}. Some closed-source commercial model~\cite{yan2025gpt-imgeval} may support the greater potential in fidelity of those methods, but haven't matched by current open-source and academic works.

\subsection{Reinforcement Learning for Diffusion Model}

For LLM, reinforcement learning has been wildely used to align with human preferences like truthfulness, helpfulness, \textit{etc.}~\cite{rlhf}.
Since the distribution formulation difference between the autoregressive model and diffusion model, the success of reinforcement learning in language models can't directly transfer to image generation models.
Prior works~\cite{black2024training, refl, xue2025dancegrpo, diffusiondpo} have explored several reinforcement learning algorithms that can be used to improve diffusion models and align with external rewards.
But most of them are focusing on text alignment and aesthetics, few works for identity similarity improvement with reinforcement learning.

\section{Methodology}
\label{sec:method}

\begin{figure*}[t]
    \centering
    \includegraphics[width=\textwidth]{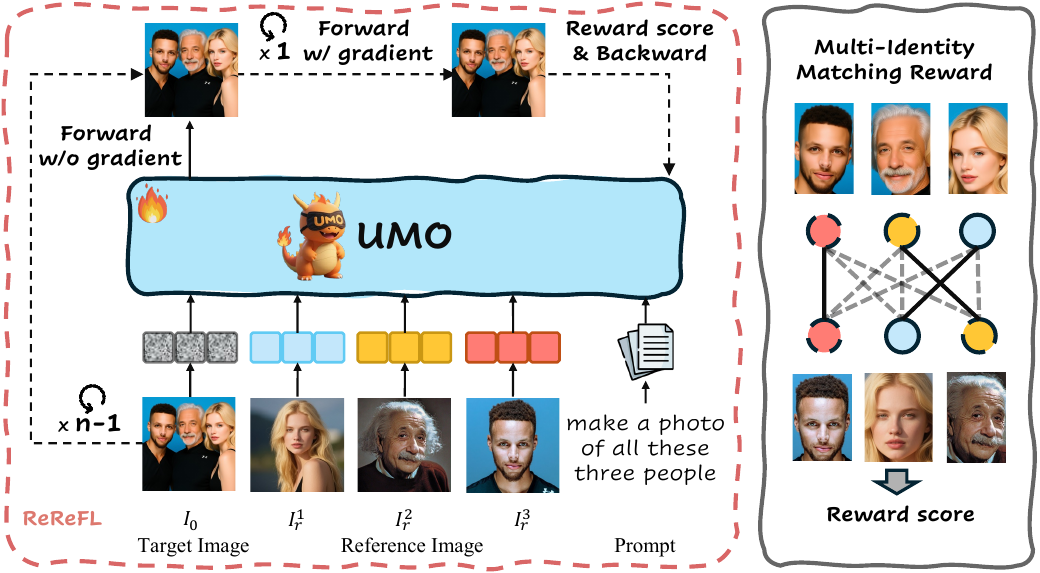}
    \caption{Illustration of the training framework of UMO. UMO's training process follows ReReFL in \Cref{alg:rerefl} with Multi-Identity Matching Reward.}
    \label{fig:umo}
\end{figure*}

\subsection{Preliminary}
\label{3.1}
Diffusion models~\cite{ddpm} have shown great capability in text-to-image generation, which sampling desired data distribution through iterative denoising. To align the models with human preferences, RLHF has been widely adapted, \textit{e.g.}, ReFL~\cite{refl} with objective
\begin{equation}
    \mathcal{L}_{\text{reward}}=\mathbb{E}_{y_i\sim \mathcal{Y}}(\phi(r(y_i,g_\theta(y_i))))
\end{equation}
where $\mathcal{Y}=\{y_1,y_2,\dots,y_n\}$ is prompt set, $\phi$ is reward-to-loss map function, $r$ is reward model and $g_\theta$ represents the generated image of diffusion model with parameters $\theta$ corresponding to prompt $y_i$.

\subsection{Data Curation Pipeline}\label{3.2}

In this paper, we aim to scale the number of identities in multi-identity preservation. But most public datasets, like X2I-subject~\cite{omnigen} and Openstory~\cite{openstory}, have few samples with identitiesx larger than two, limiting our methods. To address that, we build a data pipeline inspired by MovieGen~\cite{moviegen}, which uses the frame with multiple identities as query and recalls each identity from frames with other clips in the same long video.

Besides, we explored a synthetic data pipeline suggested by previous works such as UNO~\cite{uno} to construct a multi-identity preservation dataset. However, the identity similarity of synthetic data is relatively low. We applied strict face similarity filtering and retained only data in highly imaginative and partially stylized scenes, using them as a complement.

\begin{algorithm}[H]
    \caption{Reference Reward Feedback Learning~(ReReFL) for diffusion models}
    \label{alg:rerefl}
    \begin{algorithmic}[1] 
        \REQUIRE Image customization diffusion models $v$ with pretrained parameters $\theta_0$, pretrain loss function $\phi$, pretrain loss weight $\lambda$; reward function $R$; the number of noise scheduler time steps $T$, time step range for ReReFL finetuning $[T_s,T_e]$; dataset $\mathcal{D}=\{(y_i,I_{0_i},I^1_{r_i}, I^2_{r_i}, \dots, I^M_{r_i}) \mid i=1,2,\dots\}$ where $y_i$ is prompt, $I_{0_i}$ is target image and $I^1_{r_i}, I^2_{r_i}, \dots, I^M_{r_i}$ are reference images.
        
        \FOR{$(y_i,I_{0_i},I^1_{r_i}, I^2_{r_i}, \dots, I^M_{r_i})\in\mathcal{D}$}
            \STATE $\mathcal{L}_{\text{diff}} \gets \phi_{\theta_i}(y_i,I^1_{r_i}, I^2_{r_i}, \dots, I^M_{r_i}; I_{0_i})$ // Calculate pretrained loss with reference images
            \STATE $t \gets \mathcal{U}(T_s,T_e)$ // Pick a random time step
            \STATE $x_T \sim \mathcal{N}(0, I)$
            \FOR{$\tau=T,\dots,t+1$}
                \STATE $x_{\tau-1}\gets \text{no\_grad}(v_{\theta_i}(y_i,I^1_{r_i}, I^2_{r_i}, \dots, I^M_{r_i},x_{\tau}))$
            \ENDFOR
            \STATE $x_{t-1}\gets v_{\theta_i}(y_i,I^1_{r_i}, I^2_{r_i}, \dots, I^M_{r_i},x_t)$
            \STATE $\hat{I}_{0_i}\gets x_{t-1}$ // Predict original image by noise scheduler
            \STATE $\mathcal{L}_{\text{ReReFL}}\gets -R(\hat{I}_{0_i}; y_i,I^1_{r_i}, I^2_{r_i}, \dots, I^M_{r_i})$ // Calculate ReReFL loss with negative reward
            \STATE $\mathcal{L}\gets \lambda\mathcal{L}_{\text{diff}} + \mathcal{L}_{\text{ReReFL}}$
            \STATE $\theta_{i+1}\gets\theta_i$ // Update parameters with $\mathcal{L}$
        \ENDFOR
    \end{algorithmic}
\end{algorithm}

\begin{figure}[H]
    \centering
    \begin{subfigure}[b]{0.49\linewidth}
        \centering
        \includegraphics[width=\linewidth]{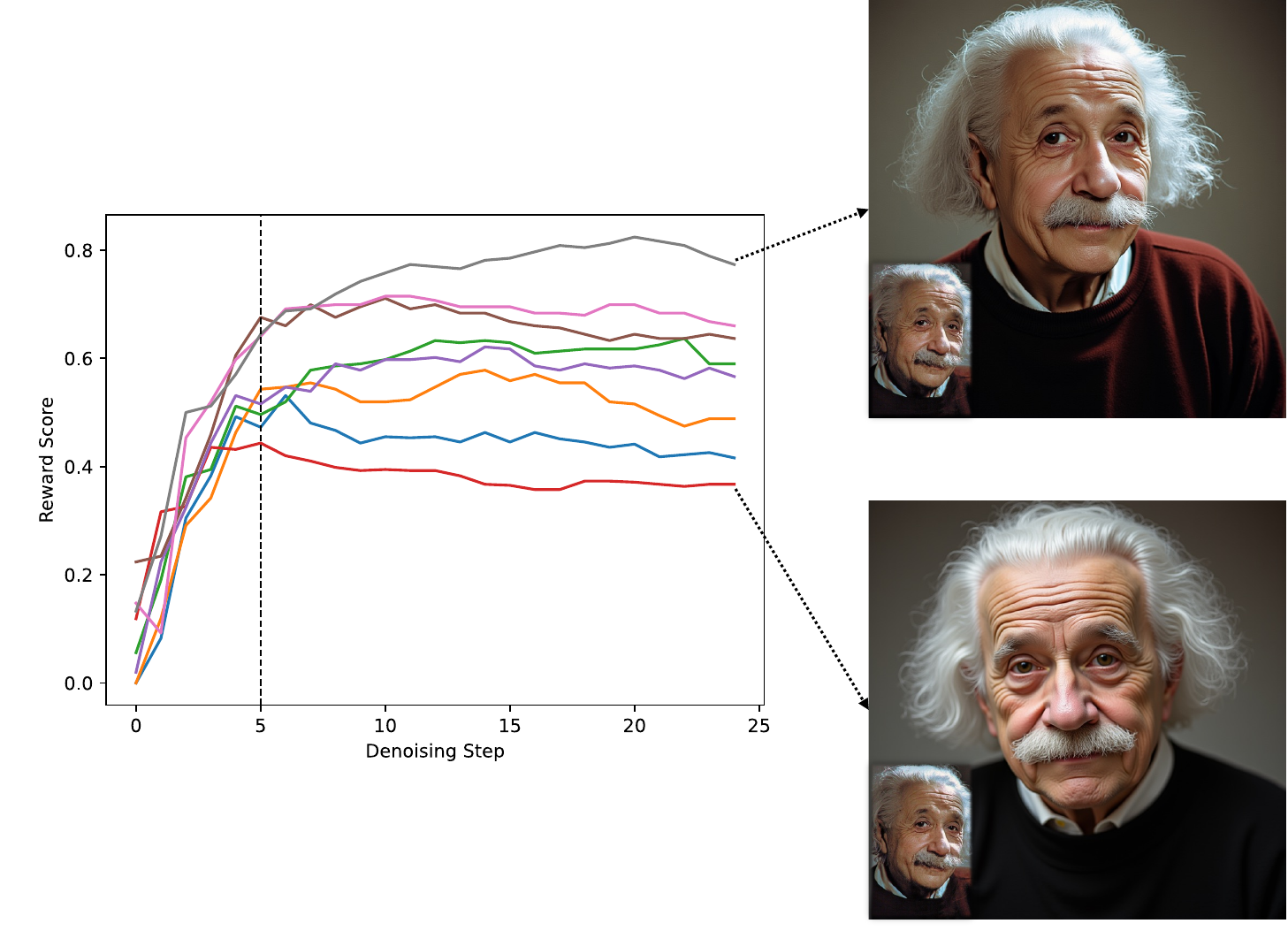}
        \caption{SIR scores of UNO~\cite{uno}.}
        \label{fig:refl_ingsight_uno}
    \end{subfigure}
    \hfill
    \begin{subfigure}[b]{0.49\linewidth}
        \centering
        \includegraphics[width=\linewidth]{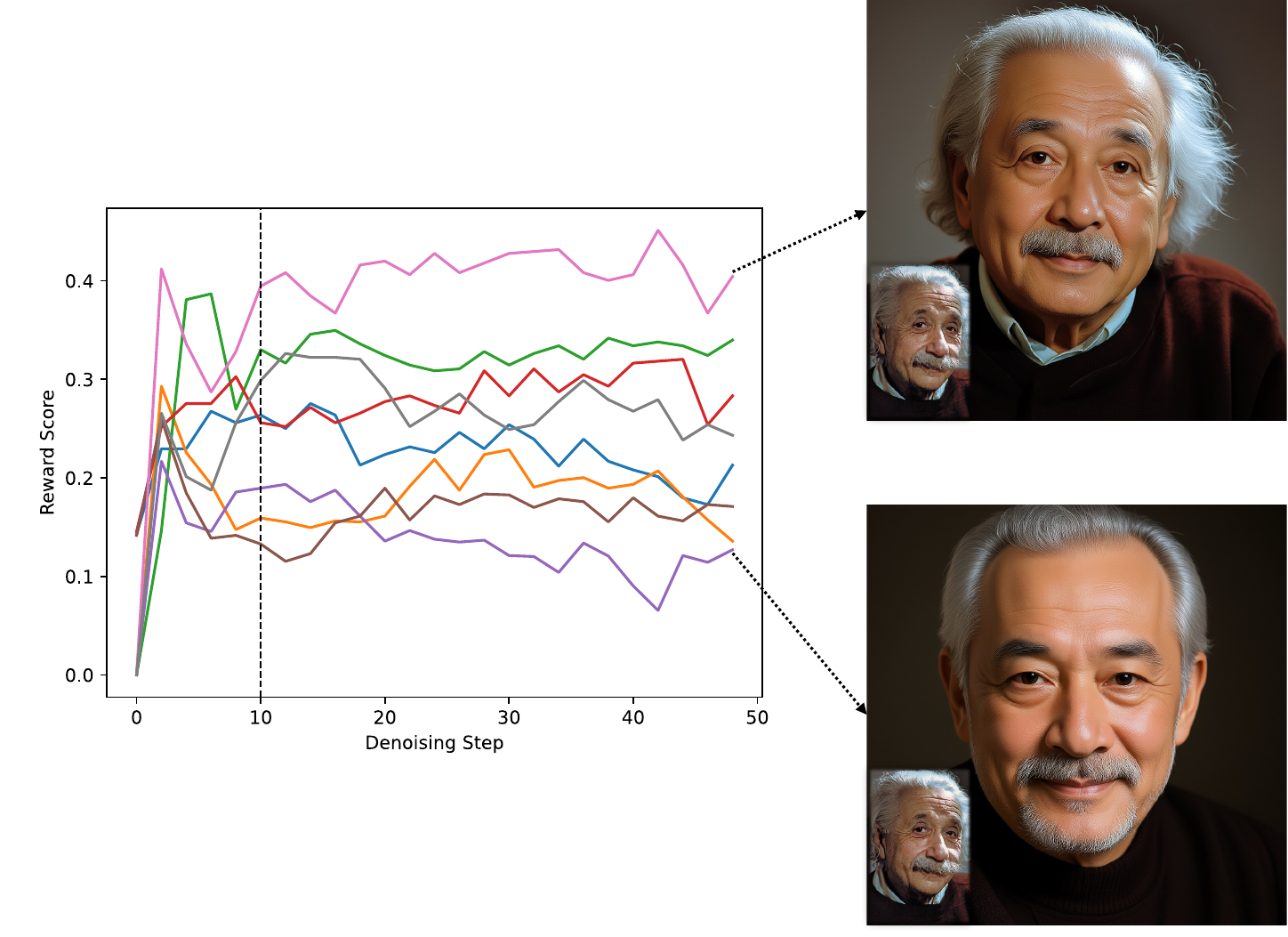}
        \caption{SIR scores of OmniGen2~\cite{omnigen2}.}
        \label{fig:refl_insight_omnigen2}
    \end{subfigure}
    \caption{Single identity reward (SIR) scores of UNO~\cite{uno} and OmniGen2~\cite{omnigen2} with different generation seeds along denoising steps. The scores become stable after step 5 and 10 respectively. And the results with highest and lowest reward scores indicating its discriminatory ability.}
    \label{fig:refl_insight}
\end{figure}

The combination of diverse data sources and the carefully designed extraction pipeline ensures the resulting multi-person identity preservation dataset features a larger number of distinct individuals and great variations beyond the identity features.

\subsection{UMO Training Framework}\label{3.3}
\subsubsection{Reinforcement Learning on Image Customization}
A trivial solution to scale multi-identity preservation is finetuning existing image customization methods with data constructed above which only achieves minor improvement as shown in \Cref{tab:ablation_uno_xverse} partly due to its relatively small proportion in diffusion models' objective. 

To steer models to align to human sensitivity towards faces, we extend ReFL to customization. Specifically, we propose Reference Reward Feedback Learning~(\textbf{ReReFL}), as illustrated in the red dashed box of \Cref{fig:umo}, which directly backpropagate reward signals with reference to the inference, as \Cref{alg:rerefl} shown.

\subsubsection{Reward with Single Identity Reference}
Effective reward is crucial to the improvement brought by RLHF. In the easiest case with only one reference identity, we introduce Single Identity Reward~(\textbf{SIR}), the cosine distance between identity embeddings to ensure a high degree of identity fidelity
\begin{equation}
    R_{\text{SIR}}=\cos(\psi(\hat{I}_{0}),\psi(I^1_{r}))
\end{equation}
where $\psi$ represents a tiny network to recognition face and get face embedding. As shown in \Cref{fig:refl_insight}, single identity reward varies drastically during former steps while getting relatively stable during latter steps. The generation result with the highest reward score preserves identity better than the one with the lowest reward score, indicating that it could serve as a reliable reward function in ReReFL.


\subsubsection{Reward with Multi-Identity Reference}
When scaling SIR to a more complicated case with multi-reference, a new challenge arises, that is improving identity fidelity while enlarging inter-ID distinction to alleviate confusion at the same time. As shown in \Cref{fig:teaser}, when identity confusion occurs, the customization models usually ignore some references, or generate a person with face from one identity while cloth from another identity.

We suggest that the keypoint lies in finding the corresponding face for each reference identity. Inspired by DETR~\cite{detr} and Multi-Object Tracking~\cite{zhang2022bytetrack}, we formulate the problem as the assignment problem under \textit{multi-to-multi matching} paradigm. Specifically, let us denote by $M$ the number of reference identities, $N$ the number of detected faces in $\hat{I}_{0}$, \textit{i.e.}, $\psi(\hat{I}_0)\in\mathbb{R}^{N\times d}$ where $d$ represents the dimension of face embedding. Each face is a vertex in a bipartite graph with one part $\hat{\mathcal{F}}$ containing all $N$ faces detected in $\hat{I}_0$ and the other part $\mathcal{F}$ containing all $M$ faces from $M$ reference identities, as illustrated in the black box of \Cref{fig:umo}. The edges are weighted by SIR of two vertices, \textit{i.e.}, $e_{\mathcal{F}_j,\hat{\mathcal{F}}_k}=\cos( \psi(\hat{I}_0)_{j}, \psi( I_r^k ) )$.
To find a maximum weight matching, we search assignment $\hat{\sigma}$ in all the potential ones $\mathfrak{S}_{n}$ of $M$ reference faces to $N$ faces in $\hat{I}_0$ with the lowest cost:
\begin{equation}
    \hat{\sigma} = \underset{\sigma \in \mathfrak{S}_{n}}{\arg\min} \sum_{i=1}^{n} \mathcal{L}_{\text{match}}(\mathcal{F}_i, \hat{\mathcal{F}}_{\sigma(i)})
    =\underset{\sigma \in \mathfrak{S}_{n}}{\arg\max} \sum_{i=1}^{n} e_{{\mathcal{F}_i}, \hat{\mathcal{F}}_{\sigma(i)}}
\end{equation}
where $\mathcal{L}_{\text{match}}(\mathcal{F}_i, \hat{\mathcal{F}}_{\sigma(i)})=-e_{{\mathcal{F}_i}, \hat{\mathcal{F}}_{\sigma(i)}}$ is a pair-wise matching cost between a reference identity $\mathcal{F}_i$ and a generated one with index $\sigma(i)$. This optimal assignment is computed efficiently with the Hungarian algorithm~\cite{hungarian}.

With the optimal assignment $\hat{\sigma}$, shown as solid lines in the black box of \Cref{fig:umo}, we find the association between reference identities and generated identities. To improve multi-identity fidelity and alleviate confusion together, we define Multi-Identity Matching Reward~(\textbf{MIMR}) as:
\begin{equation}
\label{eq:reward_ids}
    R_{\text{MIMR}}=\dfrac{1}{MN}\sum_{j=1}^N\sum_{k=1}^M (\lambda_1 \mathds{1}_{\{ k=\hat{\sigma}(j) \}} + \lambda_2 \mathds{1}_{\{ k\neq\hat{\sigma}(j) \}}) e_{\mathcal{F}_j,\hat{\mathcal{F}}_k}
\end{equation}
where $\lambda_1>0,\lambda_2<0$, adjusting gradient orientation according to $\hat\sigma$.


\section{Experiments}
\label{sec:exp}

\subsection{Experiments Setting}

\begin{figure*}[t]
    \centering
    \includegraphics[width=\linewidth]{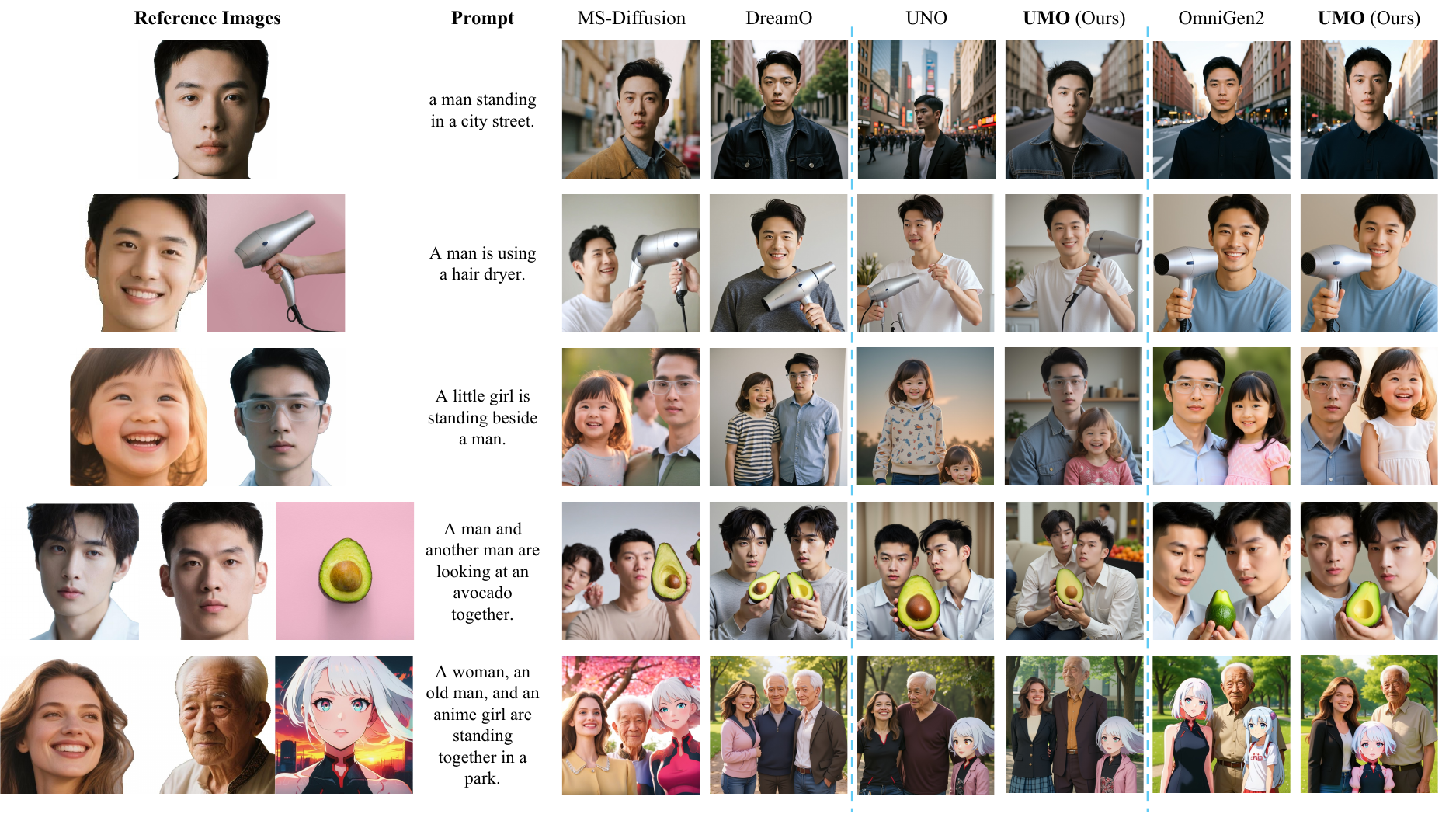}
    \caption{Qualitative comparison with different methods on XVerseBench~\cite{xverse}.}
    \label{fig:qualitative_xverse}
\end{figure*}

\begin{figure*}[h]
    \centering
    \includegraphics[width=\linewidth]{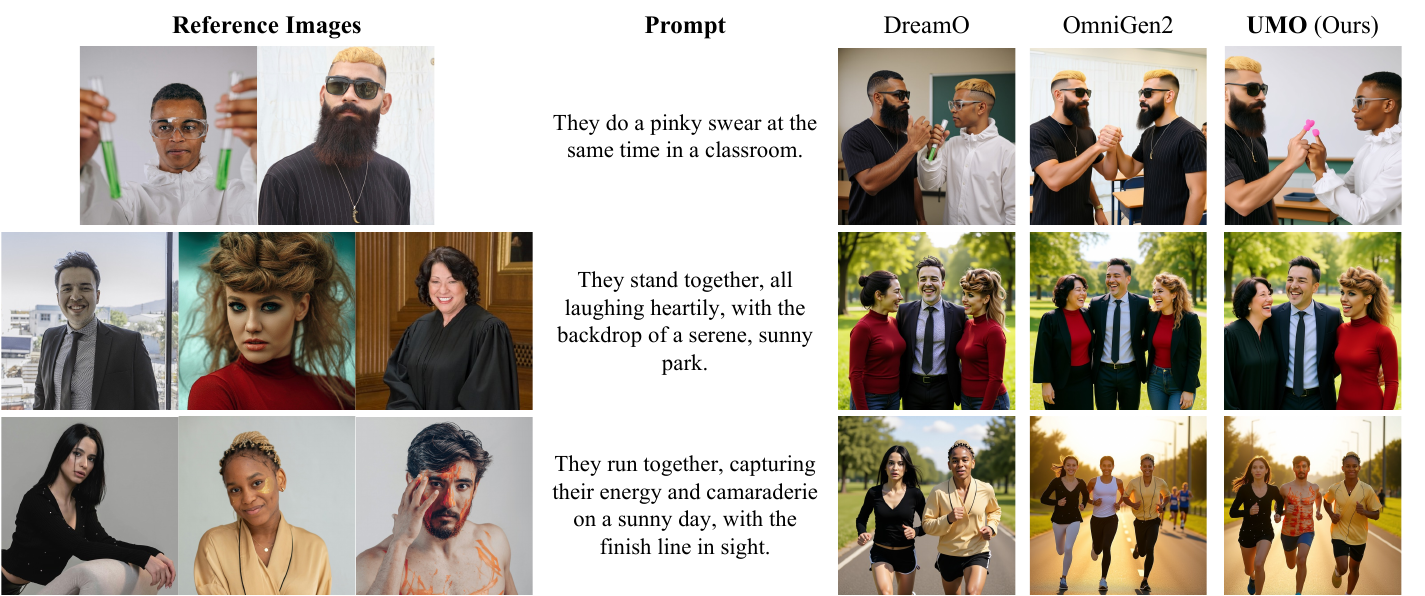}
    \caption{Qualitative comparison with different methods on OmniContext~\cite{omnigen2}.}
    \label{fig:qualitative_omnicontext}
\end{figure*}

\subsubsection{Implementation Details}
To demonstrate our UMO's generalist, the experiments are conducted based on two kinds of SOTA methods as the pretrained models:
\begin{itemize}
    \item UNO~\cite{uno}, an image customization method that supports multi-reference images based on in-context learning~\cite{iclora,omnicontrol} on DiT~\cite{dit}.
    \item OmniGen2~\cite{omnigen2}, a unified model capable of understanding and diverse generation tasks, including image editing and in-context generation.
\end{itemize}

For the training hyperparameters, we set pretrain loss weight $\lambda=1$ in \Cref{alg:rerefl} and $\lambda_1=1$, $\lambda_2=-1$ in \Cref{eq:reward_ids}. The inference-related parameters in \Cref{alg:rerefl} are set according to the suggestions of the pretrained models. Specifically, for UNO, we set the number of noise scheduler time steps $T=25$, time step range $[T_S,T_e]=[1,10]$ since the reward scores have become stable as shown in \Cref{fig:refl_ingsight_uno}, while we set $T=50$ and $[T_S,T_e]=[1,20]$ for OmniGen2 according to \Cref{fig:refl_insight_omnigen2}.

We train these models with a learning rate of $5\times 10^{-6}$ and a total batch size of $8$ on 8 NVIDIA A100 GPUs. We use LoRA~\cite{lora} with rank of $512$ during the training process. The remaining hyperparameters follow their own original settings.

\subsubsection{Comparative Methods}
UMO is a unified multi-identity optimization framework to improving identity fidelity and reducing confusion. We compare it with the two pretrained models UNO~\cite{uno} and OmniGen2~\cite{omnigen2} as baselines. Except two baselines, we compare UMO with some leading methods which support multi-reference images, including MS-Diffusion~\cite{msdiffusion}, MIP-Adapter~\cite{mip_adapter}, OmniGen~\cite{omnigen}, DreamO~\cite{dreamo} and XVerse~\cite{xverse}.

\subsubsection{Evaluation Benchmark and Metrics}
We evaluate these methods on XVerseBench~\cite{xverse} and OmniContext~\cite{omnigen2}, which cover scenarios of both single reference and multi-reference. Since OmniContext only employs GPT-4.1~\cite{gpt4.1} to assess the generated outputs, we supplement the evaluation with the ID-Sim score in XVerseBench.

To measure the severity of multi-identity confusion, we propose a new metric \textbf{ID-Conf}, which is defined as the relative margin between the two most similar generated candidate faces for a reference identity, based on the observation the confusion occurs with several indistinct generated faces. Given several reference identities $\mathcal{F}=\{\mathcal{F}_1, \mathcal{F}_2, \dots, \mathcal{F}_n\}$ and detected faces $\hat{\mathcal{F}}=\{\hat{\mathcal{F}}_1, \hat{\mathcal{F}}_2, \dots, \hat{\mathcal{F}}_m\}$ from the generated result, we define the ID-Conf metric as below:
\begin{align}
    j_i^{[1]} &\coloneqq \underset{1 \leq j \leq m}{\arg\max} \cos\big( \Psi( \mathcal{F}_i ), \Psi( \hat{\mathcal{F}_j} ) \big) \notag \\
    j_i^{[2]} &\coloneqq \underset{1 \leq j \leq m, j \neq j_i^{[1]}}{\arg\max} \cos\big( \Psi( \mathcal{F}_i ), \Psi( \hat{\mathcal{F}_j} ) \big) \\
    \text{ID-Conf} &= \dfrac{1}{n}\sum_{1 \leq i \leq n} \mathrm{clip}\bigg(
    1-\dfrac{\cos\big( \Psi( \mathcal{F}_i ), \Psi( \hat{\mathcal{F}}_{j_i^{[2]}} ) \big)} {\cos\big( \Psi( \mathcal{F}_i ), \Psi( \hat{\mathcal{F}}_{j_i^{[1]}} ) \big)}, 0, 1\bigg) \notag
\end{align}
where $\Psi$ is the model used in XVerseBench to get face embeddings. A larger value of the ID-Conf metric indicates a lower severity of identity confusion.

\subsection{Quantitative Evaluation}
We compare UMO on XVerseBench~\cite{xverse} against the two pretrained models (\textit{i.e.}, UNO~\cite{uno} and OmniGen2~\cite{omnigen2}) and other SOTA customization methods, as shown in \Cref{tab:xversebench_single} and \Cref{tab:xversebench_multi}. In both single-subject and multi-subject scenarios, UMO significantly improves the ID-Sim and ID-Conf on both pretrained models, and has a remarkable lead over previous methods, indicating the effectiveness and generalization of UMO training framework. We also evaluate UMO on OmniContext~\cite{omnigen2} as shown in \Cref{tab:omnicontext} where UMO also substantially boosts the identity consistency and alleviates confusion.

\subsection{Qualitative Analysis}



We compare with various SOTA methods on XVerseBench~\cite{xverse} to verify the effectiveness of UMO as shown in \Cref{fig:qualitative_xverse}. From the top row to the bottom one, the identity promoting effect of UMO is scalable from single identity to multi-identity scenarios and has generalization on both UNO~\cite{uno} and OmniGen2~\cite{omnigen2}. Specifically, UMO improves identity fidelity with more similar generated faces with the reference on all scenarios and alleviates confusion as the last three rows shown with more distinct generated faces with each other. For example, UNO itself gets two similar girls, both different with the reference one, in the third row while the two reference identities can be easily discriminated in the result of UMO. Also, the customization ability of general subjects is boosted or retained as the second and the fourth rows shown. As a contrast, MS-Diffusion~\cite{msdiffusion}, UNO~\cite{uno} and OmniGen2~\cite{omnigen2} all suffer low identity fidelity. Although DreamO~\cite{dreamo} shows moderate identity similarity, confusion issue occurs when the number of reference increases. Results of these models show the limited identity scalability of \textit{one-to-one mapping} paradigm, while UMO indicates the superiority of
\begin{table}[H]
    \centering
    \begin{tabular}{l|ccc}
        \toprule
        Method & ID-Sim & IP-Sim & AVG \\
        \midrule
        MS-Diffusion~\cite{msdiffusion}$^*$ & 44.12 & 76.48 & 60.30 \\
        MIP-Adapter~\cite{mip_adapter} & 39.59 & 71.97 & 55.78 \\
        OmniGen~\cite{omnigen} & 76.51 & 78.46 & 77.49 \\
        DreamO~\cite{dreamo} & 75.48 & 70.84 & 73.16 \\
        XVerse~\cite{xverse} & 79.48 & 76.86 & 78.17 \\
        \midrule
        \midrule
        UNO~\cite{uno} & 47.91 & \textbf{80.40} & 64.16 \\
        \rowcolor{myblue} \textbf{UMO}~(Ours) & \underline{80.89} & 77.09 & \underline{78.99} \\
        \midrule
        OmniGen2~\cite{omnigen2} & 62.41 & 74.08 & 68.25 \\
        \rowcolor{myblue} \textbf{UMO}~(Ours) & \textbf{91.57} & \underline{79.74} & \textbf{85.66} \\
        \bottomrule
    \end{tabular}
    \caption{Quantitative results on task type Single-Subject from XVerseBench. $^*$: We evaluate MS-Diffusion with boxes set as in MS-Bench~\cite{msdiffusion}. We highlight the \textbf{best} and the \underline{second-best} values for each metric.}
    \label{tab:xversebench_single}
\end{table}
\begin{table}[H]
    \centering
    \begin{tabular}{l|cccc}
        \toprule
        Method & ID-Sim & ID-Conf$^{\dag}$ & IP-Sim & AVG \\
        \midrule
        MS-Diffusion~\cite{msdiffusion}$^*$ & 38.98 & 66.40 & 70.98 & 58.79 \\
        DreamO~\cite{dreamo} & 50.24 & 68.67 & 64.63 & 61.18 \\
        XVerse~\cite{xverse} & 66.59 & 72.44 & \underline{71.48} & 70.17 \\
        \midrule
        \midrule
        UNO~\cite{uno} & 31.82 & 61.06 & 67.00 & 53.29 \\
        \rowcolor{myblue} \textbf{UMO}~(Ours) & \underline{69.09} & \textbf{78.06} & 68.57 & \underline{71.91} \\
        \midrule
        OmniGen2~\cite{omnigen2} & 40.81 & 62.02 & 67.15 & 56.66 \\
        \rowcolor{myblue} \textbf{UMO}~(Ours) & \textbf{71.59} & \underline{77.74} & \textbf{73.80} & \textbf{74.38} \\
        \bottomrule
    \end{tabular}
    \caption{Quantitative results on task type Multi-Subject from XVerseBench. $^{\dag}$: We align ID-Conf score with the value range of the other metrics.}
    \label{tab:xversebench_multi}
\end{table}
\begin{table}[H]
    \centering
    \begin{tabular}{l|cccc}
        \toprule
        Method & Overall & ID-Sim$^{\dag}$ & ID-Conf$^{\dag}$ & AVG \\
        \midrule
        MS-Diffusion~\cite{msdiffusion}$^*$ & 4.72 & 2.32 & 6.59 & 4.54 \\
        DreamO~\cite{dreamo} & 6.25 & 4.44 & 6.23 & 5.64 \\
        \midrule
        \midrule
        UNO~\cite{uno} & 4.71 & 1.91 & 4.91 & 3.84 \\
        \rowcolor{myblue} \textbf{UMO}~(Ours) & 5.34 & \underline{4.62} & \underline{6.60} & 5.52 \\
        \midrule
        OmniGen2~\cite{omnigen2} & \textbf{7.18} & 3.51 & 6.35 & \underline{5.68} \\
        \rowcolor{myblue} \textbf{UMO}~(Ours) & \underline{7.16} & \textbf{7.07} & \textbf{7.60} & \textbf{7.28} \\
        \bottomrule
    \end{tabular}
    \caption{Quantitative results on OmniContext. The Overall score is the geometric mean of Prompt Following (PF) and Subject Consistency (SC) scores.}
    \label{tab:omnicontext}
\end{table}
\textit{multi-to-multi matching} paradigm. Further comparison on OmniContext~\cite{omnigen2} demonstrates the positive impact on mitigating identity confusion as shown in \Cref{fig:qualitative_omnicontext}, where DreamO and OmniGen2 both suffer confusion issue, \textit{e.g.}, reference identities missing in the first row of OmniGen2 and the third row of DreamO, mismatching characteristics of each identity like mismatching hair in the first row of DreamO and wrongly assigned clothes in the second row of OmniGen2.

\begin{figure}[t]
    \centering
    \includegraphics[width=0.7\linewidth]{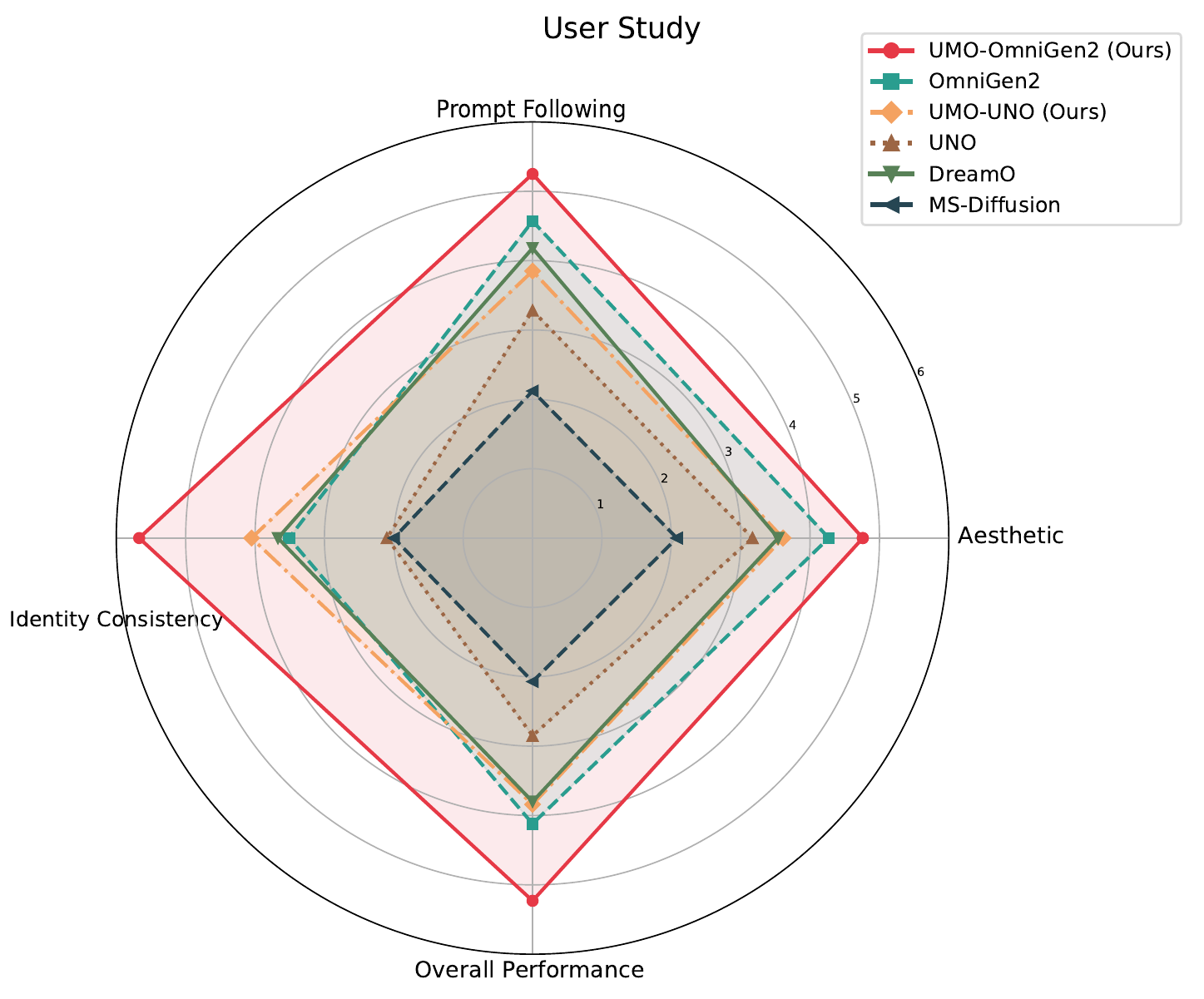}
    \caption{Radar charts of user evaluation of methods on different dimensions.}
    \label{fig:user_study}
\end{figure}

\subsection{User Study}
We further conduct a user study questionnaire to show the superiority of UMO. Questionnaires are distributed to both domain experts and non-experts, who rank the results from each method along with several dimension, \textit{i.e.}, identity consistency, prompt following, aesthetic and overall performance. As shown in \Cref{fig:user_study}, UMO achieves the best preference, demonstrating the effectiveness of \textit{multi-to-multi matching} paradigm and showcasing its capability to deliver state-of-the-art results. Also, compared to the two baselines, \textit{i.e.}, UNO~\cite{uno} and OmniGen2~\cite{omnigen2}, UMO gets significant improvement across all evaluated dimensions.

\subsection{Ablation Study}\label{subsec:ablation_study}
We conduct ablation study with UMO trained on UNO~\cite{uno} on XVerseBench~\cite{xverse} as shown in \Cref{tab:ablation_uno_xverse} and \Cref{fig:ablation} to demonstrate the effect of ReReFL and MIMR. We also conduct further ablation study with UMO trained on OmniGen2~\cite{omnigen2} on OmniContext~\cite{omnigen2} in \Cref{tab:ablation_omnigen2_omnicontext}.

\subsubsection{Effect of ReReFL}
\begin{figure}[H]
    \centering
    \includegraphics[width=0.85\linewidth]{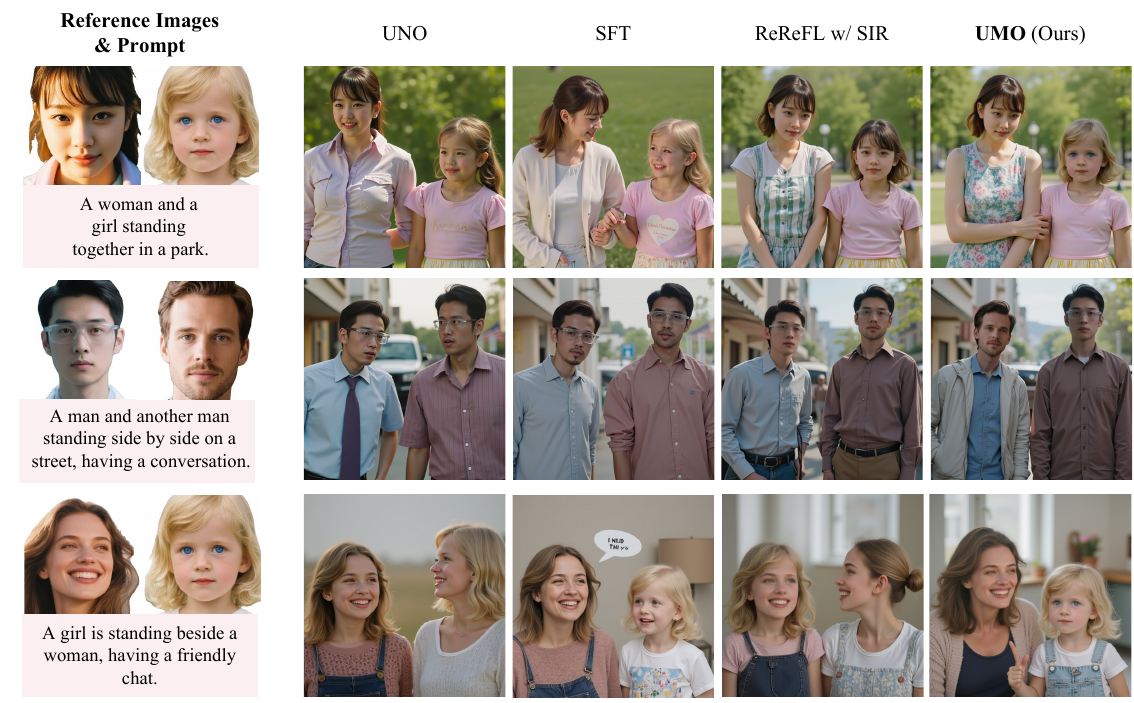}
    \caption{Visualization of ablation study. Zoom in for details.}
    \label{fig:ablation}
\end{figure}

As shown in the first two columns of \Cref{fig:ablation} and the first two rows in \Cref{tab:ablation_uno_xverse}, finetuning UNO with the same data as UMO (\textit{i.e.}, raw SFT) leads to minor improvement especially in ID-Sim and ID-Conf scores, while optimizing UNO with ReReFL demonstrates significant improvement. The comparison indicates the necessity to utilize reinforcement learning with reward focusing on facial region to unleash potential in identity consistency. Instead, vanilla SFT would suppress attention of facial feature due to its small proportion. Similar comparison on OmniGen2 in \Cref{tab:ablation_omnigen2_omnicontext} shows that the effect of ReReFL is general.

\begin{table}[H]
    \centering
    \begin{tabular}{l|cccc}
        \toprule
        Method & ID-Sim & ID-Conf$^{\dag}$ & IP-Sim & AVG \\
        \midrule
        UNO~\cite{uno} & 31.82 & 61.06 & 67.00 & 53.29 \\
        SFT & 33.94 & 62.88 & 65.17 & 54.00 \\
        ReReFL w/ SIR & \underline{65.16} & \underline{65.28} & \underline{67.25} & \underline{65.90} \\
        \rowcolor{myblue} \textbf{UMO}~(Ours) & \textbf{69.09} & \textbf{78.06} & \textbf{68.57} & \textbf{71.91} \\
        \bottomrule
    \end{tabular}
    \caption{Ablation study with UNO as the pretrained model on task type Multi-Subject from XVerseBench.}
    \label{tab:ablation_uno_xverse}
\end{table}

\begin{table}[H]
    \centering
    \begin{tabular}{l|cccc}
        \toprule
        Method & Overall & ID-Sim$^{\dag}$ & ID-Conf$^{\dag}$ & AVG \\
        \midrule
        OmniGen2~\cite{omnigen2} & \underline{7.23} & 2.86 & 6.67 & 5.59 \\
        SFT & \textbf{7.24} & 3.38 & 6.80 & 5.81 \\
        ReReFL w/ SIR & 7.14 & \underline{6.44} & \underline{7.32} & \underline{6.97} \\
        \rowcolor{myblue} \textbf{UMO}~(Ours) & 7.14 & \textbf{6.61} & \textbf{9.04} & \textbf{7.60} \\
        \bottomrule
    \end{tabular}
    \caption{Ablation study with OmniGen2 as the pretrained model on task type MULTI from OmniContext.}
    \label{tab:ablation_omnigen2_omnicontext}
\end{table}

\subsubsection{Effect of MIMR}
The last two rows of \Cref{tab:ablation_uno_xverse} demonstrate that training with SIR instead of MIMR has a significant drop in both ID-Sim and ID-Conf in multi-subject scenario. As the last two columns of \Cref{fig:ablation} shown, although SIR enhances identity similarity as well, it suffers severe confusion problem, \textit{e.g.}, result of training with SIR in the first row of \Cref{fig:ablation} has two similar identities with the reference girl missing, and the result in the last row suffers mismatching characteristics of each identity, \textit{i.e.}, the hair color of the generated woman. The comparison proves the effectiveness of MIMR through assigning correct facial supervisions to boost identity consistency and mitigate confusion. Similar observation in \Cref{tab:ablation_omnigen2_omnicontext} indicates the generalization of the effect of MIMR.

\section{Conclusion}
\label{sec:Conclusion}
In this paper, we present UMO, a Unified Multi-identity Optimization framework to improve identity consistency and alleviate confusion in multi-reference scenarios, which is based on the \textit{multi-to-multi matching} paradigm through a novel Reference Reward Feedback Learning algorithm with scalable Multi-Identity Reference Reward that reformulating multi-identity generation as a global assignment optimization problem. Additionally, we develop a scalable customization dataset along with a new metric to evaluate the extent of multi-identity confusion. Extensive experiments demonstrate that UMO significantly enhances the identity preserving ability with less confusion and greater identity scalability on various customized models, setting a new state-of-the-art among open-source methods along the dimension of identity preserving.

\clearpage

\bibliographystyle{plainnat}
\bibliography{main}

\clearpage

\onecolumn
\section*{\centering\LARGE\bfseries UMO: Scaling Multi-Identity Consistency for Image Customization via Matching Reward}
\section*{\centering Appendix}
\renewcommand{\thesection}{\Alph{section}}
\renewcommand{\thesubsection}{\Alph{section}.\arabic{subsection}}

\section{Detailed Quantitative Comparisons}
We report detailed comparison on each task type from OmniContext~\cite{omnigen2} as shown in \Cref{tab:omnicontext_single}, \Cref{tab:omnicontext_multi} and \Cref{tab:omnicontext_scene}. In all SINGLE, MULTI and SCENE task types from OmniContext, UMO significantly boosts the ID-Sim and ID-Conf on both pretrained models, \textit{i.e.}, UNO~\cite{uno} and OmniGen2~\cite{omnigen2}, leading over previous methods, \textit{e.g.}, MS-Diffusion~\cite{msdiffusion} and DreamO~\cite{dreamo}. The comprehensive evaluation demonstrate the effectiveness and generalization of UMO training framework to improve identity consistency and mitigate confusion.

\begin{table}[h]
    \centering
        \vspace{5pt}
        \begin{tabular}{l|cccc}
            \toprule
            Method & Overall & ID-Sim$^{\dag}$ & ID-Conf$^{\dag}$ & AVG \\
            \midrule
            MS-Diffusion~\cite{msdiffusion} & 5.83 & 2.89 & 6.05 & 4.92 \\
            DreamO~\cite{dreamo} & 7.65 & 5.09 & 5.83 & 6.19 \\
            \midrule
            \midrule
            UNO~\cite{uno} & 6.72 & 2.11 & 4.48 & 4.44 \\
            \rowcolor{myblue} \textbf{UMO}~(Ours) & 6.77 & \underline{5.19} & \underline{7.03} & 6.33 \\
            \midrule
            OmniGen2~\cite{omnigen2} & \textbf{7.82} & 4.75 & \textbf{7.08} & \underline{6.55} \\
            \rowcolor{myblue} \textbf{UMO}~(Ours) & \underline{7.78} & \textbf{7.95} & 6.72 & \textbf{7.48} \\
            \bottomrule
        \end{tabular}
        \vspace{5pt}
    \caption{Quantitative results on task type SINGLE from OmniContext.}
    \label{tab:omnicontext_single}
\end{table}

\begin{table}[h]
    \centering
        \vspace{5pt}
        \begin{tabular}{l|cccc}
            \toprule
            Method & Overall & ID-Sim$^{\dag}$ & ID-Conf$^{\dag}$ & AVG \\
            \midrule
            MS-Diffusion~\cite{msdiffusion} & 4.75 & 2.18 & 6.97 & 4.63 \\
            DreamO~\cite{dreamo} & 7.05 & 4.21 & 7.12 & \underline{6.13} \\
            \midrule
            \midrule
            UNO~\cite{uno} & 4.48 & 1.75 & 5.23 & 3.82 \\
            \rowcolor{myblue} \textbf{UMO}~(Ours) & 5.35 & \underline{4.46} & \underline{7.20} & 5.67 \\
            \midrule
            OmniGen2~\cite{omnigen2} & \textbf{7.23} & 2.86 & 6.67 & 5.59 \\
            \rowcolor{myblue} \textbf{UMO}~(Ours) & \underline{7.14} & \textbf{6.61} & \textbf{9.04} & \textbf{7.60} \\
            \bottomrule
        \end{tabular}
        \vspace{5pt}
    \caption{Quantitative results on task type MULTI from OmniContext.}
    \label{tab:omnicontext_multi}
\end{table}

\begin{table}[h]
    \centering
        \vspace{5pt}
        \begin{tabular}{l|cccc}
            \toprule
            Method & Overall & ID-Sim$^{\dag}$ & ID-Conf$^{\dag}$ & AVG \\
            \midrule
            MS-Diffusion~\cite{msdiffusion} & 3.95 & 1.90 & \underline{6.75} & 4.20 \\
            DreamO~\cite{dreamo} & 4.52 & 4.03 & 5.74 & 4.76 \\
            \midrule
            \midrule
            UNO~\cite{uno} & 3.59 & 1.87 & 5.03 & 3.50 \\
            \rowcolor{myblue} \textbf{UMO}~(Ours) & 4.38 & \underline{4.22} & 5.58 & 4.73 \\
            \midrule
            OmniGen2~\cite{omnigen2} & \underline{6.71} & 2.91 & 5.31 & \underline{4.98} \\
            \rowcolor{myblue} \textbf{UMO}~(Ours) & \textbf{6.78} & \textbf{6.65} & \textbf{7.03} & \textbf{6.82} \\
            \bottomrule
        \end{tabular}
        \vspace{5pt}
    \caption{Quantitative results on task type SCENE from OmniContext.}
    \label{tab:omnicontext_scene}
\end{table}

\section{More Qualitative Results}
\begin{figure*}[t]
    \centering
    \includegraphics[width=\linewidth]{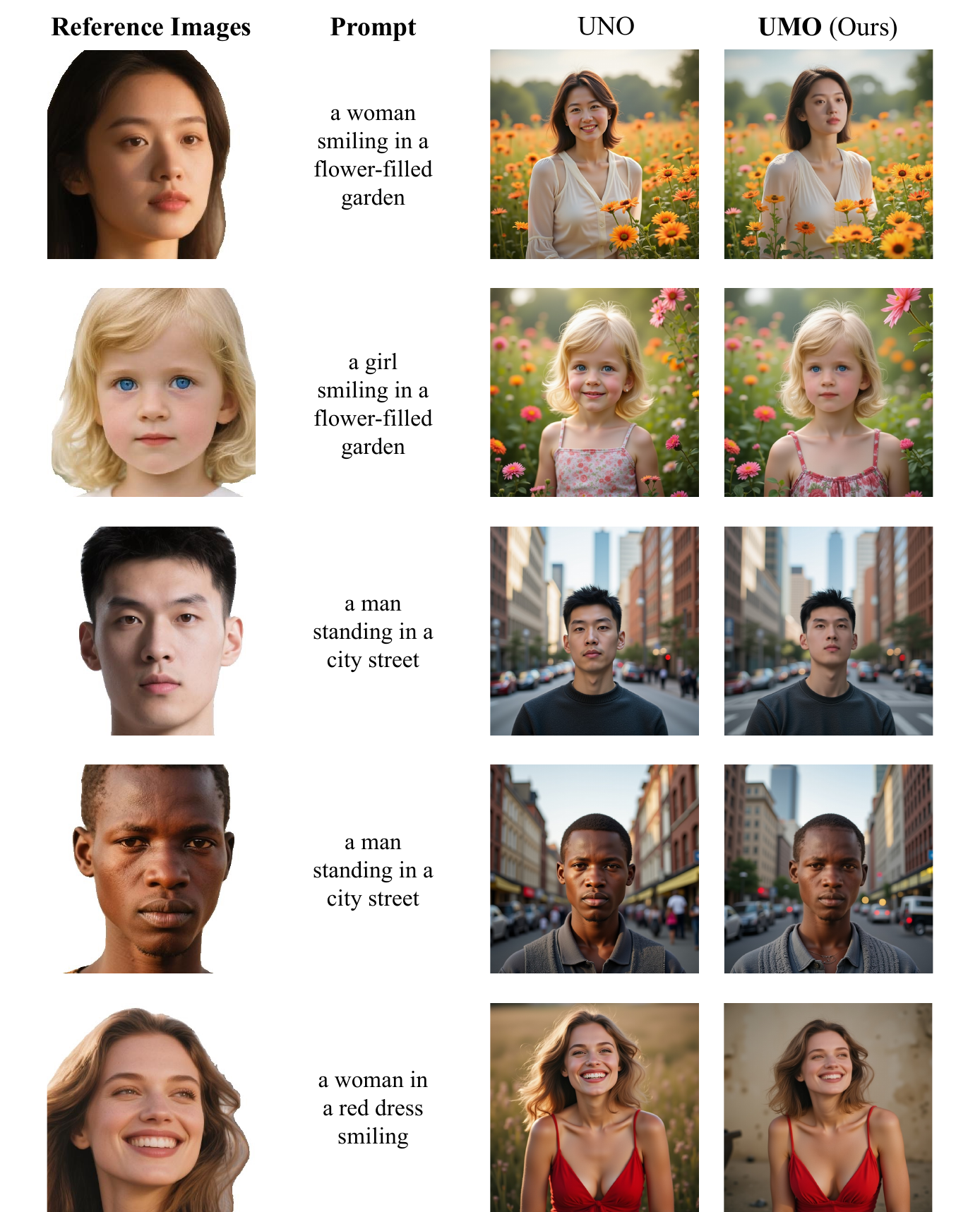}
    \caption{Qualitative results on task type Single-Subject from XVerseBench~\cite{xverse}.}
    \label{fig:supp_uno_xverse}
\end{figure*}

\begin{figure*}[t]
    \centering
    \includegraphics[width=0.85\linewidth]{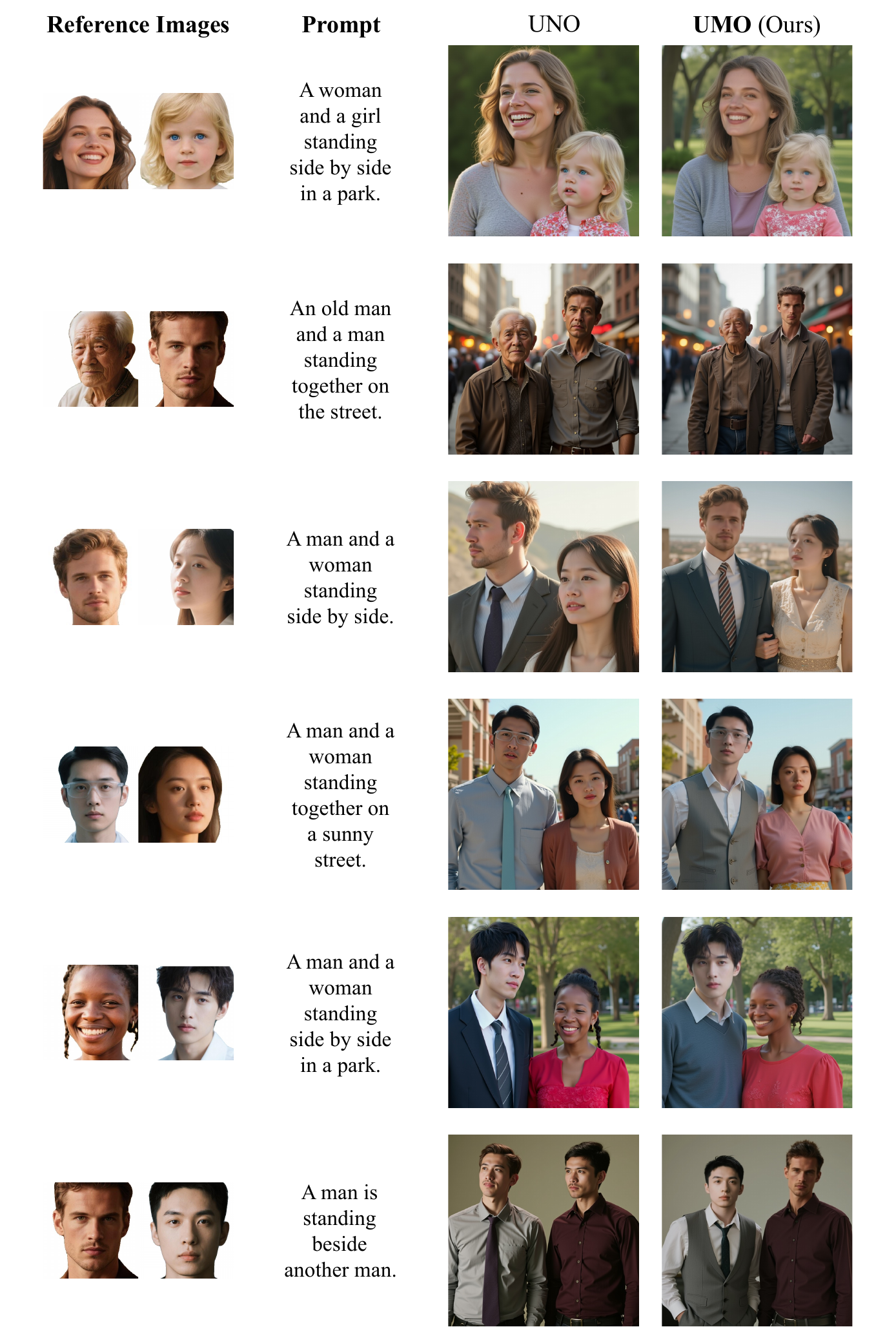}
    \caption{Qualitative results on task type Multi-Subject from XVerseBench~\cite{xverse}.}
    \label{fig:supp_uno_xverse_multi}
\end{figure*}

\begin{figure*}[t]
    \centering
    \includegraphics[width=\linewidth]{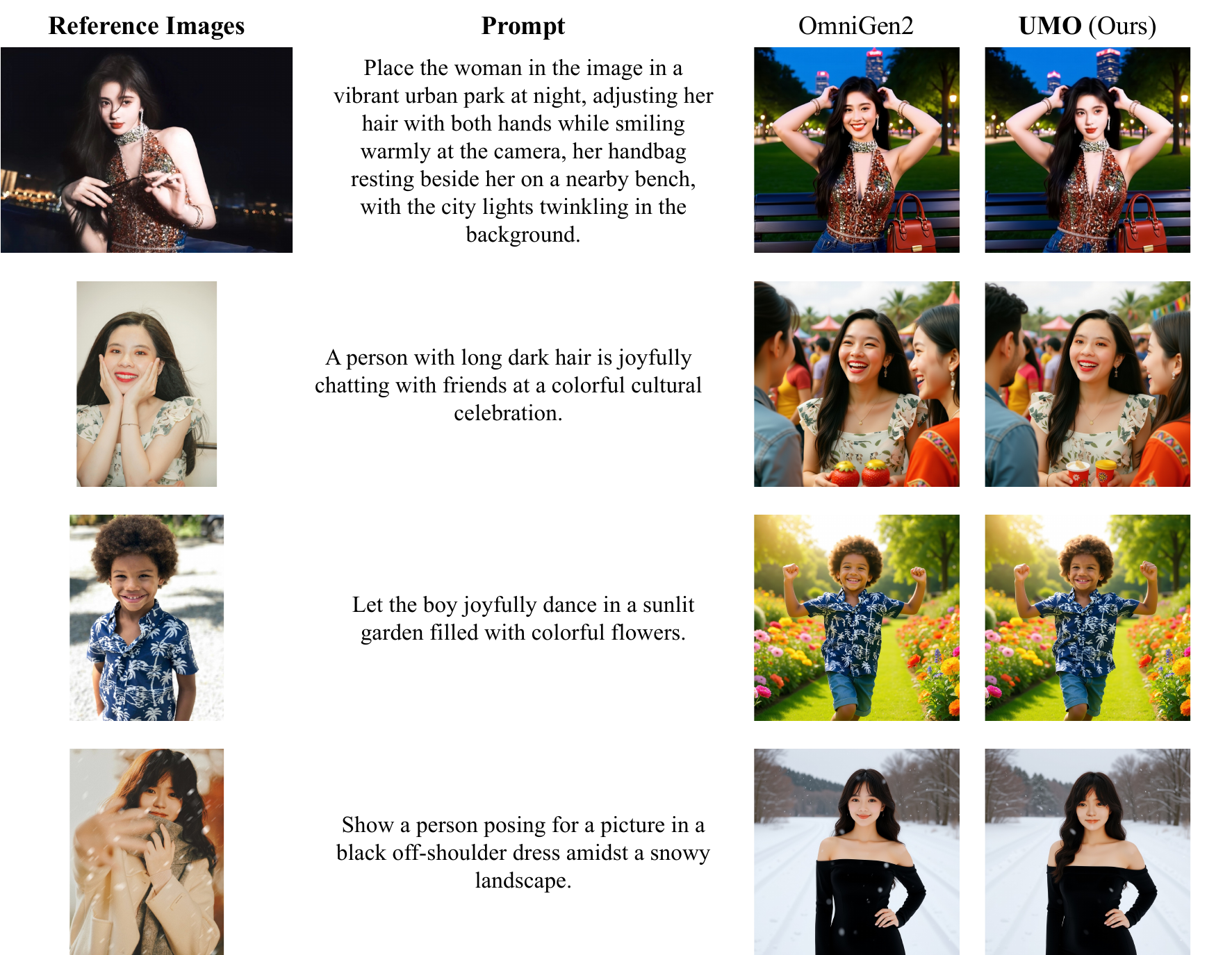}
    \caption{Qualitative results on task type SINGLE Character from OmniContext~\cite{omnigen2}.}
    \label{fig:supp_omnigen2_omnicontext}
\end{figure*}

\begin{figure*}[t]
    \centering
    \includegraphics[width=\linewidth]{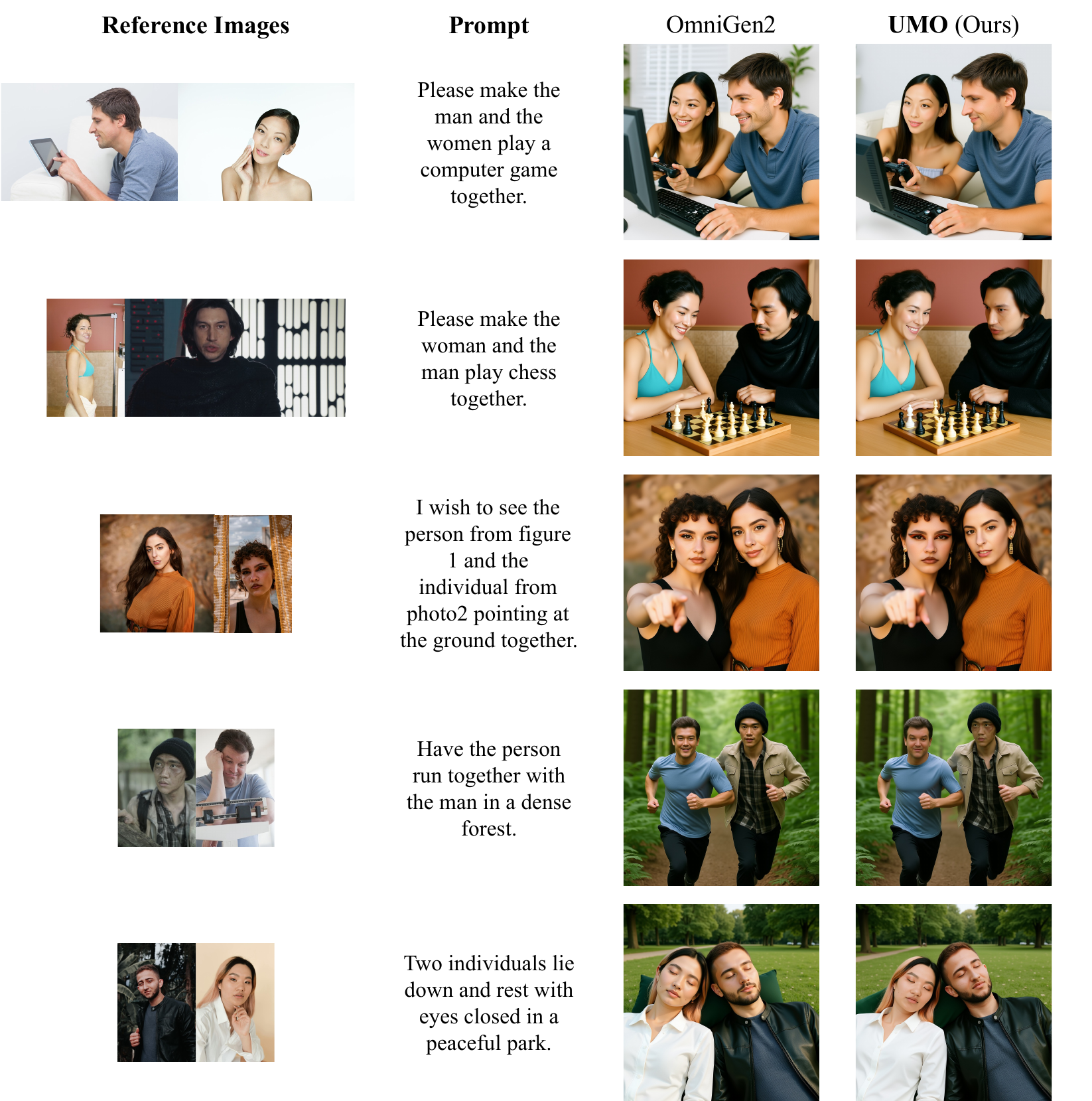}
    \caption{Qualitative results on task type MULTI Character from OmniContext~\cite{omnigen2}.}
    \label{fig:supp_omnigen2_omnicontext_multi}
\end{figure*}

We show more qualitative results on XVerseBench~\cite{xverse} in \Cref{fig:supp_uno_xverse} and \Cref{fig:supp_uno_xverse_multi}, and OmniContext~\cite{omnigen2} in \Cref{fig:supp_omnigen2_omnicontext} and \Cref{fig:supp_omnigen2_omnicontext_multi}. UMO improves identity similarity without confusion on both single identity and multi-identity scenarios, showing its general and scalable effectiveness.

\begin{itemize}
    \item \textbf{Single Identity}: In \Cref{fig:supp_uno_xverse}, UNO~\cite{uno} itself generates customization results with low identity fidelity. As a comparison, UMO gets much more similar generated identities. In \Cref{fig:supp_omnigen2_omnicontext}, although OmniGen2~\cite{omnigen2} gets moderate fidelity, UMO based on it still achieves remarkable improvement without degradation of subject similarity (\textit{e.g.}, clothes, \textit{etc}) or prompt following. The observation in single identity scenario demonstrates the extraordinary potential of UMO to enhance identity consistency across several existing models.
    
    \item \textbf{Multi-Identity}: In \Cref{fig:supp_omnigen2_omnicontext}, UNO~\cite{uno} suffers low similarity of facial features and identity confusion, \textit{e.g.}, the two generated identities in the last row have almost the same facial features which is the ``average'' facial features of the two reference ones. By contrast, UMO shows its superiority with higher fidelity and without identity confusion. In \Cref{fig:supp_omnigen2_omnicontext_multi}, the results of OmniGen2~\cite{omnigen2} show moderate identity similarity, while UMO still boosts it without degradation. The observation in the scenario of multi-identity shows the impressive promoting ability of UMO to improve identity fidelity and alleviate confusion on existing image customization methods.
\end{itemize}

\section{Discussion}
Although we build UMO to maintain high-fidelity identity preservation and alleviate identity confusion with scalability to multi-identity, stably scaling to more identities is still restricted due to the dramatic decrease of the pretrained models' reference ability when the number of reference images or identities increases, which demonstrates a similar view with~\cite{yue2025does}.

\begin{table}[h]
    \centering
    \begin{tabular}{lc}
        \toprule
        \textbf{Scenarios} & \textbf{Prompts} \\
        \midrule
        \multirow{4}{*}{Single Identity} & (1) The man on the beach. \\
         & (2) The girl holds a board written "UMO" \\
         & (3) The man is skateboarding on the street. \\
         & (4) Transform the style of image into Sketch style \\
        \midrule
        \multirow{5}{*}{Single Identity + Subject} & (1) The man holding the can. \\
         & (2) The elf wearing the T-shirt in the second image, \\
         & with "UXO Team" written on it. \\
         & (3) The woman in the first image is running \\
         & with the dog in the second image. \\
        \midrule
        \multirow{5}{*}{Two Identities} & (1) The man and the woman are sitting in a classroom \\
         & (2) The man in the first image and the man in the second image \\
         & shake hands and look straight ahead. \\
         & (3) Half-body portrait of the woman in the first image and the old man \\
         & in the second image, retro comic style \\
        \midrule
        \multirow{2}{*}{Two Identities + Subject} & (1) The woman is holding the toy, while the man standing beside. \\
         & (2) The man is holding a hair dryer and drying the woman's hair. \\
        \midrule
        \multirow{2}{*}{More Identities} & (1) The three people are playing cards. \\
         & (2) Portrait of the three people, oil painting style. \\
        \bottomrule
    \end{tabular}
    \caption{The detailed prompts used in \Cref{fig:showcase}.}
    \label{tab:showcase_prompt}
\end{table}
\end{document}